\documentclass{bmvc2k}
\usepackage{multirow}

\title{Part-based Face Recognition with Vision Transformers}

\addauthor{Zhonglin Sun}{zhonglin.sun@qmul.ac.uk}{1}
\addauthor{Georgios Tzimiropoulos}{g.tzimiropoulos@qmul.ac.uk}{1}

\addinstitution{
 School of Electronic Engineering and Computer Science\\
 Queen Mary university of London\\
 London, UK}

\runninghead{Z.Sun and G.Tzimiropoulos}{Part-based Face Recognition with fViT}

\usepackage{amsfonts}
\begin{document}

\maketitle

\begin{abstract}
Holistic methods using CNNs and margin-based losses have dominated research on face recognition. In this work, we depart from this setting in two ways: (a) we employ the Vision Transformer as an architecture for training a very strong baseline for face recognition, simply called \textit{fViT}, which already surpasses most state-of-the-art face recognition methods. (b) Secondly, we capitalize on the Transformer's inherent property to process information (visual tokens) extracted from irregular grids to devise a pipeline for face recognition which is reminiscent of part-based face recognition methods. Our pipeline, called \textit{part fViT}, simply comprises a lightweight network to predict the coordinates of facial landmarks followed by the Vision Transformer operating on patches extracted from the predicted landmarks, and it is trained end-to-end with no landmark supervision. By learning to extract discriminative patches, our part-based Transformer further boosts the accuracy of our Vision Transformer baseline achieving state-of-the-art accuracy on several face recognition benchmarks.
\end{abstract}

\section{Introduction}
\label{sec:intro}
Face recognition(FR) is an important problem in computer vision with many applications such as border control and surveillance. With the advent of Deep Learning, the de-facto pipeline for FR over the last years comprises (a) a CNN( Convolutional Neural Network) backbone, which processes the face image holistically to compute a facial feature embedding which is used to calculate a similarity score, and (b) an appropriate loss function for discriminative embedding learning. While the bulk of recent work on FR has focused on (b), i.e., designing more effective loss functions~\cite{schroff2015facenet,wen2016discriminative,liu2017sphereface,wang2018cosface,arcface,deng2021variational,li2021spherical}, this work mostly focuses on (a) i.e. devising new architectures for facial feature extraction. 

\begin{figure}[htbp]
  \centering
   \includegraphics[width=0.6\linewidth]{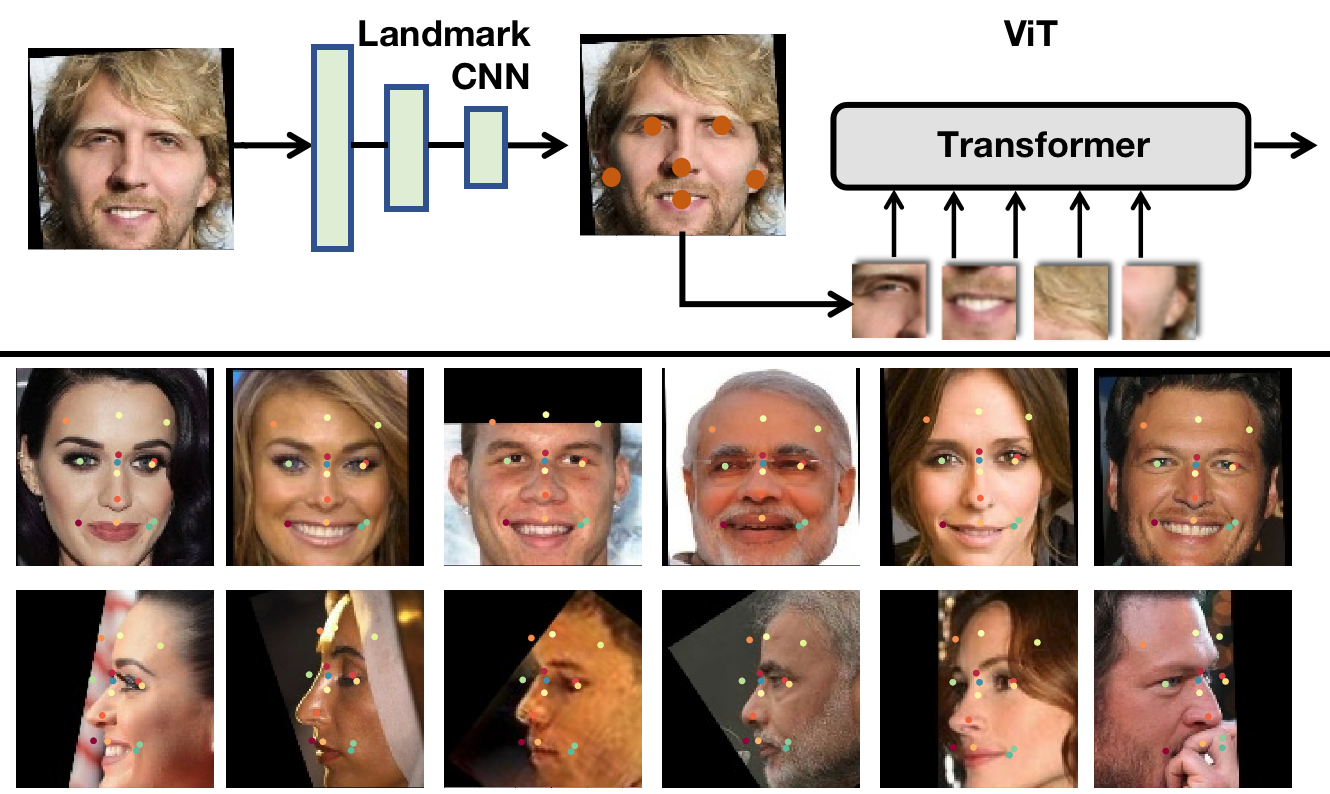}

   \caption{Illustration of our part-based ViT for face recognition. A facial image is processed by a lightweight landmark CNN which produces a set of facial landmarks. The landmarks are used to sample facial parts from the input image which are then used as input to a ViT for feature extraction and recognition. The whole system is trained end-to-end without landmark supervision. Examples of landmarks detected by the landmark CNN are shown.}
   \label{fig:Visual_landmark}
\end{figure}

The first motivation of our work is the recently introduced Vision Transformer~\cite{dosovitskiy2021an}, which is gaining increasing popularity in Computer Vision with recent results reported being very competitive to the ones produced by CNN backbones~\cite{Liu_2021_ICCV,xiao2021early}. Hence, our first contribution is to explore how far one can go with a vanilla ViT for face recognition using the vanilla loss of~\cite{wang2018cosface}. We show that such a backbone with appropriate hyper-parameter optimization already achieves state-of-the-art results for face recognition. The second motivation for our work is that the ViT, contrary to CNNs, can actually operate on patches extracted from irregular grids and does not require the uniformly spaced sampling grid used for convolutions. As the human face is a structured object composed of parts (e.g., eyes, nose, lips), and inspired by seminal work on part-based face recognition before deep learning~\cite{chen2013blessing}, in this paper, we propose to apply ViT on patches representing facial parts. Specifically, our second contribution is a newly proposed parts-based pipeline for deep face recognition where discriminatively learned landmarks are firstly predicted through a lightweight landmark CNN, patches are extracted around them and then fed to a ViT. Notably, the whole system, called part fViT, can be trained end-to-end without landmark supervision. Fig.~\ref{fig:Visual_landmark} shows an overview of the proposed pipeline. 

In summary, \textbf{our contributions} are: 
\begin{itemize}
    \item We appropriately train a vanilla ViT for face recognition using a vanilla loss, which we coin \textit{fViT}, and show that fViT produces state-of-the-art results on several popular face recognition benchmarks. 
    \item We capitalize on the Transformer architecture to propose a new pipeline for face recognition, coined \textit{part fViT}, where discriminatively learned patches are firstly extracted and then fed to the ViT for recognition, essentially building a part-based ViT for face recognition. Notably, the landmark CNN used for predicting the landmarks is trained end-to-end with the ViT without landmark supervision. 
    \item We show that our part fViT surpasses our strong baseline fViT setting a new state-of-the-art on several face recognition datasets. Moreover we ablate several components of our pipeline illustrating their impact on face recognition accuracy. 
    \item We show that the landmark CNN which is part of our pipeline, is effective for the side task of unsupervised landmark discovery.
\end{itemize}
    

\section{Related Work}

A detailed review of face recognition papers is out of scope, herein we focus on losses, Region-aware methods and Vision Transformers which are more related to our work.

\noindent \textbf{Loss functions:} Several papers~\cite{schroff2015facenet,wen2016discriminative,liu2017sphereface,wang2018cosface,arcface,deng2021variational,li2021spherical} have focused on learning features which are both separable and discriminative through using an appropriate loss function. While separability can be achieved with the softmax loss, learning discriminative features is more difficult as, within the mini-batch, training cannot see the global feature distribution~\cite{wen2016discriminative}. To this end, FaceNet~\cite{schroff2015facenet} uses triplets to directly learn a mapping to a compact Euclidean space such that facial features from the same identity are as close as possible while features from different identities are as far as possible. 

To avoid the problem of triple selection, Center loss~\cite{wen2016discriminative} minimizes the distance between the learned deep features for each face and their corresponding class centres in order to achieve intra-class concentration. Observing that the inter-class boundaries are not well separated in Softmax Loss, L-softmax~\cite{liu2016large} considers the joint formulation of softmax cross-entropy loss and linear layer, penalizing the distance of the class boundary, resulting in more discriminative features.   Following that, CosFace\cite{wang2018cosface} applied normalization not only on the weights, but also on the feature embedding, and proposed to add the margin on $cos(\theta)$ where $\theta$ is the angle between linear weight and embedding. ArcFace\cite{deng2019arcface} further defined the margin on the angle $\theta$ rather than $cos(\theta)$. VPL\cite{deng2021variational} pays attention to learning the prototype of each class by regarding the distribution of classes on the feature space, and proposed to change the static prototype by injecting memorized features for approximating the prototype variation. Recently, Sphereface2~\cite{wen2021sphereface2} proposes to conduct binary classification for recognition, and a number of general principles are also summarized in the work on how to design a good loss.

\noindent \textbf{Region-aware methods:} 
Although CNNs provide standard backbones for face recognition relying on global information, they ignore the fact that the face is a structured object with parts which can be used for more effective learning of facial features. For example, the seminal work of~\cite{chen2013blessing}, which was the state-of-the-art before the advent of deep learning, shows that extracting a very large number of multi-scale features around 5 pre-defined landmarks (e.g. eye, nose, mouth) can be very effective for face recognition. To address local features via deep learning-based solutions, TUA\cite{liu2015targeting} proposed to integrate local and global face features from different disjoint CNN via different GPUs, to aggregate the feature concatenation operation is used. FAN-Face~\cite{yang2020fan} explored how features from a pre-trained facial landmark localization network can be used to enhance face recognition accuracy, however the landmark localization and recognition networks were not jointly trained. Moreover, \cite{ding2017trunk,kang2019hierarchical,kang2018pairwise} have all come up with methods to extract landmark-related features during CNN training, however, they still require pre-defined landmarks. To avoid explicit landmark supervision, Comparator Networks~\cite{xie2018comparator} propose a pipeline that performs attention to multiple discriminative local regions (landmarks), and uses them to compare local descriptors between pairs of faces. Finally, HPD~\cite{wang2020hierarchical} takes full use of the attention mechanism to predict attention masks for local features.

Our part fViT is inspired by~\cite{chen2013blessing,xie2018comparator} but works~\textit{in a completely different manner}. Firstly, landmarks are learned by directly predicting their x,y coordinates using a very lightweight network (i.e. mobilenetV3~\cite{howard2019searching}). Then patches centred at the predicted landmarks are sampled and fed to a Transformer~\cite{vaswani2017attention,dosovitskiy2021an} for face recognition. Notably we take advantage of the Transformer architecture to provide as input a set of patches sampled at irregular spatial locations which departs from standard face recognition methods based on CNNs which use a regular image grid (necessary to define convolutions) but also from ViT~\cite{dosovitskiy2021an} which also uses a regular grid for processing an input image. Moreover, our system is trained in an end-to-end manner without landmark supervision.

\noindent \textbf{Vision Transformer:}
The Transformer was firstly introduced in Natural Language Processing for machine translation and other NLP tasks~\cite{vaswani2017attention}. It comprises Self-attention and Feed-Forward layers. Vision Transformer (ViT) was introduced in~\cite{dosovitskiy2021an}, and since then it has been shown to provide competitive accuracy to CNNs~\cite{xiao2021early}. Training ViT is more difficult compared to CNNs~\cite{touvron2021going,touvron2021training}. A number of approaches have been proposed to facilitate ViT's training~\cite{steiner2021train,wang2021pyramid,yuan2021tokens,touvron2021going,yuan2021tokens,wu2021cvt,yuan2021incorporating,Liu_2021_ICCV,xiao2021early,graham2021levit,chen2021visformer}. 
In this work, we discard the previous approach using ViT for face recognition~\cite{zhong2021face} where regular overlapped patches are extracted from faces, instead we adopted the standard ViT backbone~\cite{dosovitskiy2021an} with the training improvements of~\cite{steiner2021train}. This already gives us a very strong baseline which surpasses most existing state-of-the-art methods for face recognition on MS1M~\cite{guo2016ms} dataset. Next, we go beyond~\cite{dosovitskiy2021an} and follow-up works~\cite{steiner2021train,wang2021pyramid,yuan2021tokens,touvron2021going,yuan2021tokens,wu2021cvt,yuan2021incorporating,Liu_2021_ICCV,xiao2021early,graham2021levit,chen2021visformer} by applying the transformer, for the first time to the best of our knowledge on a set of patches extracted from \textit{non-regular grids} provided by a lightweight network which is trained end-to-end to provide discriminative landmarks without explicit supervision.





\begin{figure*}[htbp]
  \centering
   \includegraphics[width=0.68\linewidth]{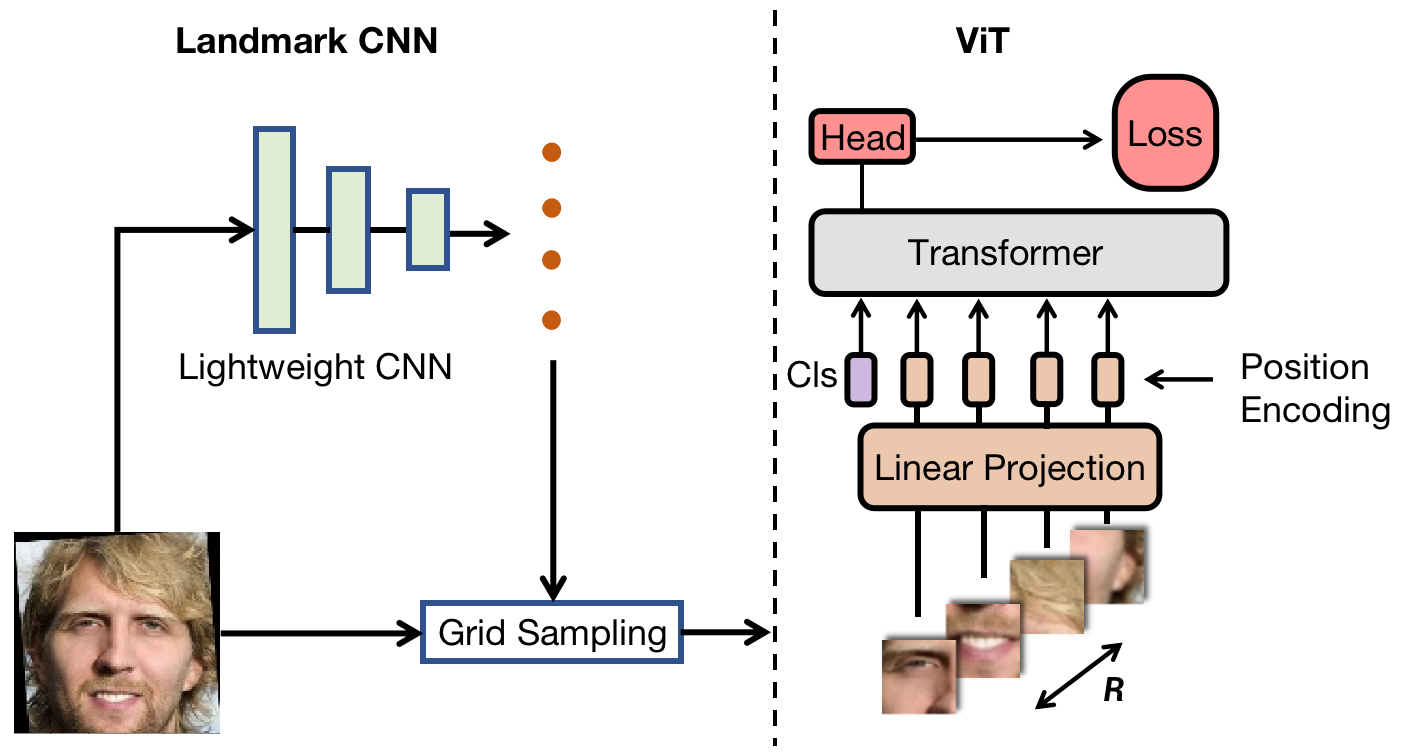}

   \caption{The overall structure of our proposed part fViT: A lightweight CNN is used to predict a set of facial landmarks. Then, differentiable grid sampling is applied to extract the discriminative facial parts which are then used as input to a ViT for feature extraction and recognition. Yellow nodes represent the regressed facial landmark coordinates extracted}
   \label{fig:Landmark_strcuture}
\end{figure*}

\section{Methodology}\label{sec:method}

In Section~\ref{sec:pure_vit}, we firstly describe our strong baseline, called fViT, obtained by training ViT with CosFace loss. Then in Section~\ref{sec:landmark_vit}, we introduce our proposed part-based ViT for face recognition, called part fViT.



\subsection{fViT: ViT for Face Recognition}\label{sec:pure_vit}

We are given a facial image $\mathbf{X}\in\mathbb{R}^{H \times W \times C }$ ($C=3$). Following ViT~\cite{dosovitskiy2021an}, the image is divided into $R=P\times P$ non-overlapping patches which are then mapped into visual tokens using a linear embedding layer $\mathbf{E}\in\mathbb{R}^{P^2 \times d}$. To preserve spatial information a positional embedding $\mathbf{p}_{s}\in\mathbb{R}^{P^{2}\times d}$ is also learned which is added to the initial visual tokens. Then, the token sequence is processed by $L$ Transformer layers.

The visual token at layer $l$ and spatial location $s$ is $\mathbf{z}^l_{s}\in\mathbb{R}^d, \;\;\; l=0,\dots,L-1, \;\; s=0,\dots,P^{2}-1.$  In addition to the $R$ visual tokens, a classification token $\mathbf{z}^l_{cls}\in\mathbb{R}^{d}$ is prepended to the token sequence~\cite{devlin2018bert}.
The $l-$th Transformer layer processes the visual tokens $\mathbf{Z}^l\in\mathbb{R}^{(P^{2}+1) \times d}$ of the previous layer using a series of  Multi-head Self-Attention (MSA), Layer Normalization (LN), and MLP ($\mathbb{R}^d \rightarrow \mathbb{R}^{4d} \rightarrow \mathbb{R}^d$) layers as follows:
\begin{eqnarray}\label{eq:transformer}
\mathbf{Y}^{l} & = & \textrm{MSA}(\textrm{LN}(\mathbf{Z}^{l-1})) + \mathbf{Z}^{l-1},\\
\mathbf{Z}^{l} & = & \textrm{MLP}(\textrm{LN}(\mathbf{Y}^{l})) + \mathbf{Y}^{l}. \label{eq:transformer2}
\end{eqnarray}
A single Self-Attention (SA) head is given by:

\begin{equation}
\mathbf{y}^{l}_{s} = \sum_{s'=0}^{P^{2}-1} \sigma\{(\mathbf{q}^{l}_{s} \cdot \mathbf{k}^{l}_{s'})/\sqrt{d_h}\} \mathbf{v}^{l}_{s'}, s=0,\dots,P^{2}-1,
\label{eq:SA}
\end{equation}

where $\sigma(.)= \textrm{Softmax}(.)$,  $\mathbf{q}^{l}_{s},\mathbf{k}^{l}_{s}, \mathbf{v}^{l}_{s} \in\mathbb{R}^{d_h}$ are the query, key, and value
vectors computed from $\mathbf{z}^l_{s}$ using embedding matrices $\mathbf{W_q},\mathbf{W_k}, \mathbf{W_v} \in\mathbb{R}^{d \times d_h}$, $d_h$ is the scale factor in self-attention. Finally, the outputs of the $h$ heads are concatenated and projected using embedding matrix $\mathbf{W_h}\in\mathbb{R}^{hd_h \times d}$.

The classification token $\mathbf{z}^L_{cls}$ is trained for face recognition using the CosFace  loss~\cite{wang2018cosface}:

\begin{equation}\label{equ:cosface}
    \mathbf{Loss}=\frac{1}{N}\sum_{i} -\textrm{log}\frac{\mathbf{e}^{\mathbf{b}(\textrm{cos}(\mathbf{\theta}_{\mathbf{y}_{\mathbf{i}},\mathbf{i}})-\mathbf{m})}}
    {\mathbf{e}^{\mathbf{b}(\textrm{cos}(\mathbf{\theta}_{\mathbf{y}_{\mathbf{i}},\mathbf{i}})-\mathbf{m})}+\sum_{j\neq \mathbf{y}_{\mathbf{i}}}\mathbf{e}^{\mathbf{b}\textrm{cos}(\mathbf{\theta}_{\mathbf{j},\mathbf{i}})}},
\end{equation}
where $N$ is the number of samples in a batch, $\mathbf{z}=\frac{\mathbf{z}^{L}_{cls}}{||\mathbf{z}^{L}_{cls}||}$, $\mathbf{z}_i$ is the $i-$th sample and $\mathbf{y}_{i}$ the corresponding ground-truth, $\mathbf{W}=\frac{\mathbf{W}^{*}}{||\mathbf{W}^{*}||}$ is the weight matrix of the last linear layer, $W_{j}$ is the normalized $j-$th column (class) of the weight matrix, $\textrm{cos}(\theta_{\mathbf{y_{i}},\mathbf{i}})= W_{\mathbf{y_{i}}}^{T}\mathbf{z}_{\mathbf{i}}$, $\mathbf{m}$ is the margin and $\mathbf{b}$ is fixed to be $||\mathbf{z}^L_{cls}||$.


We found that fViT, similarly to ViT is prone to overfitting. Hence, to obtain high accuracy, we used a combination of approaches for training including stochastic depth regularization~\cite{larsson2016fractalnet}, random resize \& crop, RandAugment~\cite{cubuk2020randaugment}, Cutout, and finally Mixup~\cite{zhang2017mixup}. The details of the choice of these are given in supplementary material $2.1$.

\subsection{Part fViT} \label{sec:landmark_vit}

The ViT as described by Eqs.~\ref{eq:transformer} \&~\ref{eq:transformer2} operates on a sequence of visual token which do not need to be computed on uniform grid. Inspired by 
work on part-based FR~\cite{chen2013blessing}, in this section we describe how to apply ViT on patches representing facial parts.

Specifically, we use a lightweight weight CNN to predict a set of $R=P\times P$ landmarks: 
\begin{equation}
\mathbf{r} = \textrm{CNN}(\mathbf{X}),\; r_i=[x_i,y_i]^T,\; i=1,\dots,P^{2},  
\end{equation}
where for our CNN we used a MobilenetV3~\cite{howard2019searching}.  

Then, we sample a patch centered at each landmark coordinate $r_i$. To accommodate for fractional coordinates, we used the differentiable grid sampling method of STN~\cite{jaderberg2015spatial} for extracting each patch. Following this, each patch is tokenized by the embedding layer $\mathbf{E}$, giving rise to $R$ part tokens which together with the class token are processed by the Transformer of Eqs.~\ref{eq:transformer} \&~\ref{eq:transformer2}. We explore a number of options for the positional encodings added to the part tokens in an ablation study in Section ~\ref{sec:ablation}. 

The whole pipeline, called part fViT is very simple, and is shown in Fig.~\ref{fig:Landmark_strcuture}. It is trained end-to-end with no landmark supervision using simply the CosFace loss of Eq.~\ref{equ:cosface}. Notably, the landmark regression network forms an information bottleneck which was previously found useful in methods for unsupervised landmark discovery~\cite{jakab2018unsupervised}. We also confirm this finding in an ablation study in Section ~\ref{sec:ablation}. Finally, although heatmap regression methods with softmax could be used, we opted for direct coordinate regression which is simpler.

\section{Experiments}\label{sec:experiments}
In this section, we evaluate accuracy of the proposed face transformers on several well-known datasets and compare them with that of recently proposed state-of-the-art methods.

\subsection{Implementation details}\label{sec:implementation}
For training, and for a fair comparison with other methods, we used the refined version~\cite{deng2019lightweight} of MS1M~\cite{guo2016ms} (MS1MV3) containing 93,431 identities unless specificed. We also provide result training on VGGFace2~\cite{cao2018vggface2} with 3.1M images and 8.6K identities. Face images are of resolution $112 \times 112$ and aligned (provided by~\cite{deng2019arcface})
We tested our models on LFW~\cite{huang2008labeled}, CFP-FP~\cite{sengupta2016frontal}, AgeDB-30~\cite{moschoglou2017agedb}, IJB-B\cite{whitelam2017iarpa}, IJB-C\cite{maze2018iarpa} and MegaFace\cite{kemelmacher2016megaface} for conducting recognition performance evaluation. For LFW, CFP-FP and AgeDB-30, we use 1:1 verification accuracy(\%). We report TAR@FAR=1e-4 results on IJB-B and IJB-C. For Megaface, Megaface/id refers to the rank-1 identification accuracy (\%) on 1M distractors, and Megaface/ver refers to TAR@FAR=1e-6 verification accuracy. For training the Transformer, we opted to use a large amount of data augmentation compared to the original 
FR setting used in ResNets, please refers to supplementary material Section $2.1.1$ and $2.1.2$ for more details regarding hyper-parameters, augmentations, model structure and training details.

\subsection{Ablation Studies}\label{sec:ablation}

We conducted a number of studies to highlight the impact of different design choices for our face Transformers. Our ablation studies are mainly carried out on the patch number $R=49$ for its efficient training speed. We also attached the \textbf{improvement of data augmentation}, \textbf{degree of overlap} and \textbf{Effect of different landmark CNNs} in the supplementary material Section $2.2$.



\paragraph{Effect of patch number and different fViT models:} Our first experiment focuses on how the number of patches (or equivalently the number of landmarks $R$ for the part fViT) impacts the accuracy of the proposed face Transformers. The number of patches chosen are 16, 49 and 196 with the FLOPs 1.17G, 3.3G and 12.64G respectively, and both fViT-B and fViT-S models are tested, as illustrated in Table~\ref{number_patches}. Note that when the number of patches increases, the patch size $K$ is reduced; specifically for 196 landmarks the corresponding patch size is $8$ and for 16 landmarks, the patch size is $28$, ensuring that for the case of small number of landmarks the whole facial image is still analyzed. fViT-B has feature dim with 768 and MLP dim with 2048 while fViT-S has 512 and MLP dim with 2560. In both cases the number of heads is 11. The results are shown in Table~\ref{number_patches}. 

A number of interesting conclusions can be drawn by this experiment: \textbf{(1)} More patches (landmarks) result in more accurate prediction, as expected. \textbf{(2)} When the number of patches (landmarks) is very large (i.e. 196) then the part fViT outperforms fViT by small margin. \textbf{(3)} As the number of patches/landmarks decreases this gap increases specifically for CFP-FP and AgeDB. This is important as models processing fewer tokens are significantly more lightweight. For example the 49 landmark model is $4\times$ faster than the 196 landmark model. 
\begin{table}[ht]
\centering
\footnotesize
\begin{tabular}{|c|c|c|c|c|c|c|}
\hline
Backbone & Patch No. & Model & LFW & CFP-FP & AgeDB& IJB-C \\ \hline
\multirow{6}{*}{fViT-B} & 196 & part fViT & 99.83 & \textbf{99.21} & \textbf{98.29}& \textbf{97.29}\\ \cline{2-7} 
 & 196 & fViT & 99.85 & 99.01 & 98.13& 97.21\\ \cline{2-7} 
 & 49 & part fViT & 99.80 & 98.78 & 97.85&96.37 \\ \cline{2-7} 
 & 49 & fViT & 99.78 & 98.00 & 97.56& 96.30 \\ \cline{2-7} 
 & 16 & part fViT & 99.80 & 97.30 & 97.22& 94.90 \\ \cline{2-7} 
 & 16 & fViT & 99.78 & 96.87 & 96.46 &94.85\\ \cline{2-7} 
 \hline \hline
\multirow{6}{*}{fViT-S} & 196 & part fViT & 99.83 & 99.09 & 98.18& 96.58\\ \cline{2-7} 
 & 196 & fViT & 99.83 & 98.90 & 97.90&96.50\\ \cline{2-7} 
 & 49 & part fViT & 99.80 & 98.7 & 97.81 &96.33\\ \cline{2-7} 
 & 49 & fViT & 99.80 & 98.0 & 97.31 &96.05\\ \cline{2-7} 
 & 16 & part fViT & 99.71 & 97.25 & 97.06 &94.21\\ \cline{2-7}
 & 16 & fViT & 99.71 & 96.95 & 96.25 &94.19\\ \hline
\end{tabular}
\caption{Impact of number of patches and different fViT models on FR accuracy.}
\label{number_patches}
\end{table}

\paragraph{Effect of different positional encodings}\label{sec:pos_enc}  Herein, we explore the function of positional encoding in our part fViT-B $R=49$ landmarks. We test 3 types of positional encodings: (a) trainable ones as in the original fViT~\cite{dosovitskiy2021an}, (b) cosine~\cite{vaswani2017attention} and (c) coordinate-based. For coordinate-based, we used a linear layer to embed each landmark $r_i$ into $\mathcal{R}^d$ and then added this vector to the corresponding visual token. Results are shown in Table~\ref{tab:Multi_experiments} (top section). As it can be observed the trainable one and the coordinate-based achieve the best accuracy. 

\begin{table}[h]
\centering
\footnotesize
\scalebox{0.9}{
\begin{tabular}{|c|c|c|c|c|c|}
\hline
Experiment & Content & LFW & CFP-FP & AgeDB & IJB-C \\ \hline
\multirow{3}{*}{Positional encoding} &Trainable & 99.80 & \textbf{98.78} & 97.85 & \textbf{96.37} \\ \cline{2-6} 
 & Cosine & 99.80 & 98.65 & \textbf{98.03} & 96.08 \\ \cline{2-6} 
 & Coordinate & 99.80 & 98.71 & 97.66 & 96.29 \\ \hline \hline
 \multirow{2}{*}{Information bottleneck } & w/ IB & 99.80 & \textbf{98.78} & \textbf{97.85} & \textbf{96.37} \\ \cline{2-6} 
  &  w/o IB & 99.76 & 97.73 & 97.31 & 96.05 \\ \hline \hline
\multirow{4}{*}{Unsupervised landmark} & Vanilla fViT  & 99.78 & 98.00 & 97.56 & 96.30\\ \cline{2-6}
 & part fViT (MobilenetV3)  & 99.80 & \textbf{98.78} & \textbf{97.85} & \textbf{96.37} \\ \cline{2-6} 
& part fViT (FAN (Frozen)) & 99.36 & 95.31 & 96.11 & 93.96 \\ \cline{2-6} 
 & part fViT (MobilenetV3 (Frozen)) & 99.81 &98.72 & 97.66 & 96.35 \\ \hline
\end{tabular}}
\caption{Results of various ablation studies: (a) Top section: impact of different positional encodings. (b) Middle section: impact of information bottleneck. (c) Last section: impact of unsupervised landmark discovery. All experiments are with part fViT-B with $R=49$.}
\label{tab:Multi_experiments}
\end{table}
\paragraph{Effect of information bottleneck:}\label{sec:bottleneck} We experimented with providing to the part fViT as input the penultimate layer's feature from the landmark CNN, essentially injecting features from the CNN to the fViT and violating the information bottleneck of our pipeline in Section~\ref{sec:landmark_vit}. Specifically, the CNN penultimate layer's feature was concatenated with the (trainable) positional encoding and then projected to $\mathcal{R}^d$. Results are shown in Table~\ref{tab:Multi_experiments} (middle section). As observed, violating the information bottleneck leads to decreased accuracy.

\paragraph{Effect of unsupervised landmark discovery:} Since supervised facial landmark localization methods are widely used in literature, we compare our part fViT with a model that uses the landmarks provided by a state-of-the-art facial landmark localization, namely FAN~\cite{bulat2017far}. We freeze the landmark CNN part from the well-trained part fViT to train a new ViT, coined as part fViT(mobilenet (Frozen)). Results are shown in Table~\ref{tab:Multi_experiments} (bottom section). As it can be observed, using FAN (Pretrained  and frozen parameters) to provide the input landmarks to fViT reduces to suboptimal performance. This way of directly using patches of landmarks provided by an accurate supervised landmark network leads to worse results than training a vanilla fViT. With a pretrained R=49 landmark network and only training the fViT part, we achieved a significant improvement than FAN network.  We can conclude that for directly using patches of landmarks on the FR task, FAN is unable to provide the proper landmarks.


\begin{table}[h]
\centering
\footnotesize
\scalebox{0.9}{
\begin{tabular}{|c|c|c|c|c|c|c|c|}
\hline
Method & LFW & CFP-FP & AgeDB  & IJB-B & IJB-C & MegaFace/id & MegaFace/ver \\ \hline
CosFace\cite{wang2018cosface} & 99.81 & 98.12 & 98.11  & 94.80 & 96.37 & 97.91 & 97.91  \\ \hline
ArcFace\cite{deng2019arcface}& 99.83 & 92.27 & 92.28 & 94.25 & 96.03 & 98.35 & 98.48 \\ \hline
GroupFace\cite{kim2020groupface}& 98.85 & 98.63 & 96.20 & 94.93 & 96.26 & 98.74 & 98.79  \\ \hline
CircleLoss\cite{sun2020circle}& 99.73 & 96.02 & - &  - & 93.95 & 98.50 & 98.73  \\ \hline
DUL\cite{chang2020data}& 99.83 & 98.78 & - & - & 94.61 & 98.60 & -  \\ \hline
CurricularFace\cite{huang2020curricularface}& 99.80 & 98.37 & 98.32  & 94.8 & 96.1 & 98.71 & 98.64 \\ \hline
Sub-center ArcFace\cite{deng2020sub}& 99.80 & 98.80 & 98.31 & 94.94 & 96.28 & 98.16 & 98.36 \\ \hline
FAN-Face\cite{yang2020fan}& 99.85 & 98.63 & 98.38 &  94.97 & 96.38 & 98.70 & 98.95 \\ \hline
BroadFace\cite{kim2020broadface}& 99.85 & 98.63 & 98.38 &  94.97 & 96.38 & 98.70 & 98.95 \\ \hline
ArcFace-challenge\cite{deng2021masked}& 99.85 & 99.06 & 98.48 & - & 96.81 & - & - \\ \hline
VPL\cite{deng2021variational}& 99.83 & 99.11 & \textbf{98.60} &  95.56 & 96.76 & 98.80 & \textbf{98.97}  \\ \hline
ALN\cite{zhang2021adaptive}& - & 96.53 & 97.25  & 93.13 & 95.27 & - & -  \\ \hline
VirFace\cite{li2021virtual}& 99.56 & 97.15 & - &  88.90 & 90.54 & - & -  \\ \hline
MagFace\cite{meng2021magface}& 99.83 & 98.46 &  96.15 & 94.51 & 95.97 & - & -  \\ \hline
SCL\cite{li2021spherical}& 99.80 & 98.59 & 98.26 &  94.74 & 96.09 & 81.40 & 97.15  \\ \hline
Face Transformer~\cite{zhong2021face}  & 99.83 & 96.19 & 97.82 &  - & 95.96 & - & -  \\ \hline\hline
fViT-B, ours & \textbf{99.85} & 99.01 & 98.13  & 95.97 & 97.21 & 98.69 & 98.91  \\ \hline
Part fViT-B, ours & 99.83 & \textbf{99.21} & 98.29  & \textbf{96.11} & \textbf{97.29} & \textbf{98.96} & 98.78  \\ \hline
\end{tabular}}
\caption{Comparison with the state-of-the-art on multiple datasets. Our baseline fViT and part fViT achieve state-of-the-art results on most datasets.}
\label{SOTA}
\end{table}

\subsection{Comparison with the State-of-the-Art}

We chose our part fViT-B and fViT with patch size 8 and R=196 to compare with recently proposed state-of-the-art FR methods. The landmark CNN used was MobilenetV3.


\paragraph{Quantitative results:} We report the results of the models trained on MS1MV3, and tested on various benchmarks. The results are shown in Table~\ref{SOTA}. As observed, on LFW which is saturated, our proposed methods achieved top accuracy along with a few other methods. On the pose-sensitive dataset CFP-FP, our part-fViT has obtained the accuracy of 99.21\%, surpassing the other state-of-the-art methods of VPL~\cite{deng2021variational} and Arcface-challenge\cite{deng2021masked}. Similar results are observed for IJB-B and IJB-C benchmarks: not only does our part fViT outperform the other state-of-the-art methods by significant margin (97.29 TAR on IJB-C, 96.11 TAR on IJB-B), but even our baseline fViT is the second best method (97.21 TAR on IJB-C and 95.97 TAR on IJB-B). Similar results are obtained on MegaFace evaluation, where our part fViT is the top performing along with a few other methods. The only exception is on AgeDB-30, where our part fViT obtains 98.29\%. We need to mention that the loss function used is CosFace~\cite{wang2018cosface} which was chosen for its simplicity and stability. It is possible that using more advanced loss functions for training, including VPL~\cite{deng2021variational}, ArcFace~\cite{deng2019arcface} and Sphereface2~\cite{wen2021sphereface2}.
\begin{table}[h]
\centering
\footnotesize
\scalebox{0.9}{
\begin{tabular}{|c|c|c|c|c|c|c|}
\hline
                & LFW   & AgeDB-30 & IJB-B & IJB-C & MegaFace/Id & MegaFace/Ver \\ \hline
Comparator Networks~\cite{xie2018comparator} & -     & -        & 85.0  & 88.5  &         &          \\ \hline
FAN-Face~\cite{yang2020fan}        & -     & -        & 91.1  & \textbf{93.5}  & -       & -        \\ \hline
SphereFace~\cite{liu2017sphereface}      & 99.55 & 92.88    & 89.41 & 91.96 & 71.53   & 85.02    \\ \hline
CosFace~\cite{wang2018cosface}         & 99.51 & 92.98    & 88.61 & 90.98 & 71.65   & 85.45    \\ \hline
ArcFace~\cite{deng2019arcface}        & 99.47 & 91.97    & 89.11 & 91.60 & 73.65   & 87.77    \\ \hline
Circle Loss~\cite{sun2020circle}     & 99.48 & 92.90    & 88.56 & 90.83 & 71.32   & 84.34    \\ \hline
SphereFace2~\cite{wen2021sphereface2}     & 99.50 & 93.68    & \textbf{91.31 }& 93.25 & \textbf{74.38}   & \textbf{89.19}    \\ \hline \hline
fViT, Ours            & 99.44 & 93.52    & 88.13  & 90.26 & 71.11   & 85.04    \\ \hline
part fViT, Ours       & \textbf{99.56} & \textbf{93.92}    & 88.98 & 91.03 & 71.63   & 85.91    \\ \hline
\end{tabular}}
\caption{Comparison with the state-of-the-art results on VGGFace2.}
\label{VGG2}
\end{table}
We also conducted experiments on the VGGFace2 dataset using similar parameters with Resnet64 in SphereFace~\cite{liu2017sphereface} to show the results of our part fViT in Table~\ref{VGG2}. Despite adding a large amount of data augmentation, our baseline fViT perform worse than the results provided by Resnet64 which is similar to the Face Transformer when training on a small scale dataset such as CASIA-webface~\cite{yi2014learning}. Our part fViT also achieves a better result than the baseline fViT when training on MS1M, while it is still a little worse than the Resnet64 with advanced losses(e.g. ArcFace~\cite{deng2019arcface}). Our future work will investigate how our method works on other large scale benchmarks like Glink360~\cite{an2021partial}.

\begin{figure*}[ht]
  \centering
  \includegraphics[width=0.9\linewidth]{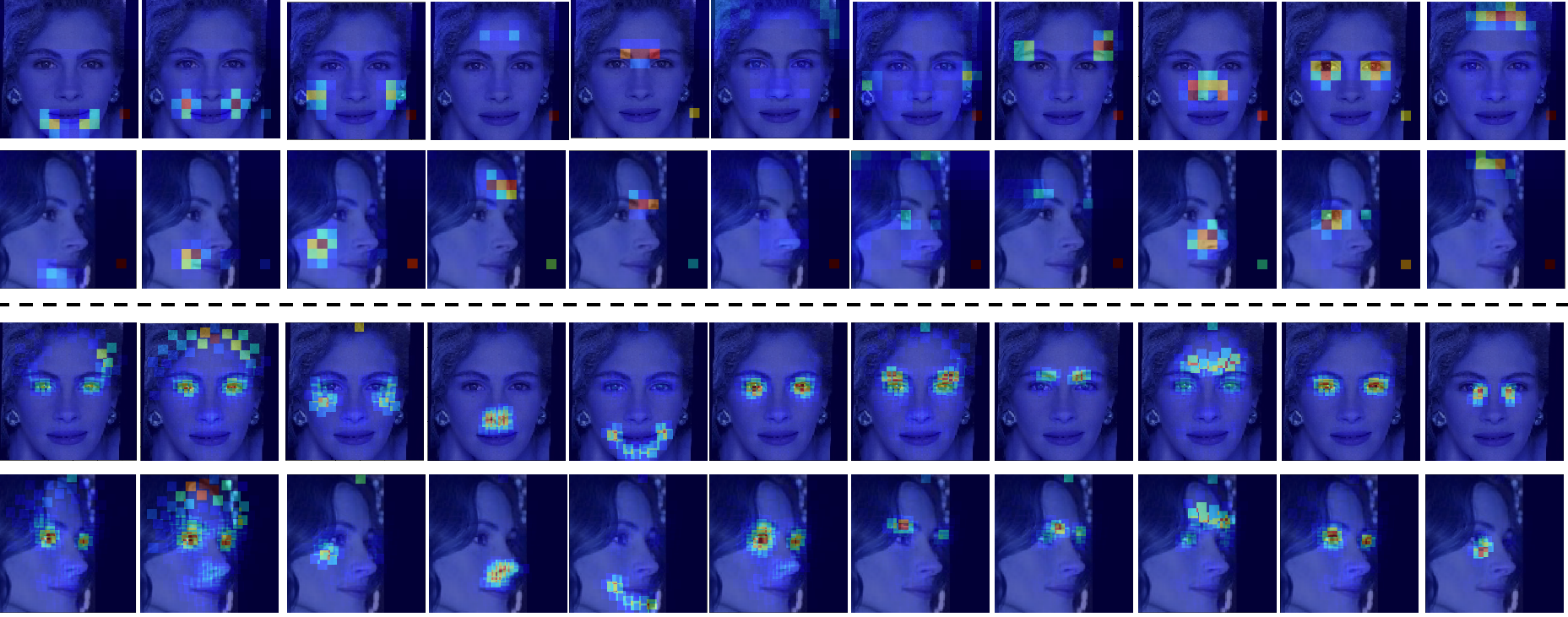}
  \caption{Visualization of attention maps. The first and second rows show the 11 attention maps produced by the 11 heads of the baseline fViT-B; The third and fourth rows show the 11 attention maps produced by the 11 heads of the part fViT-B with $R=196$ landmarks.
  }
  \label{fig:attention_map}
\end{figure*}

\begin{figure*}[htbp]
  \centering
  \includegraphics[width=0.95\linewidth]{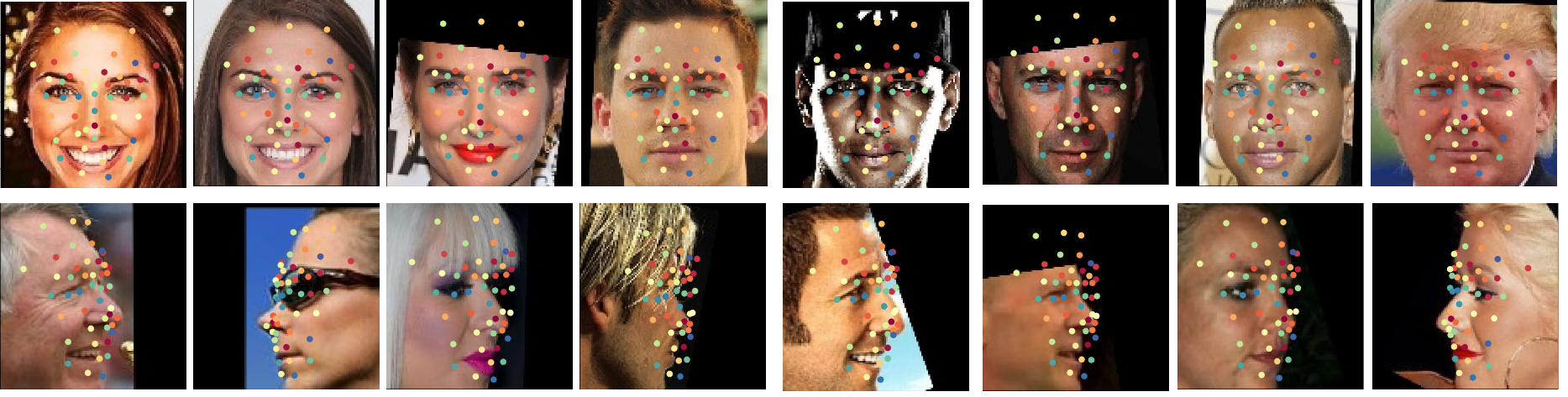}

  \caption{Visualization of the learned landmarks from our part fViT-B with $R=49$. landmarks of same colour in different images across pose was learned to some good degree.}
  \label{fig:49landmark}
\end{figure*}
\paragraph{Qualitative results:} We first compare the attention maps produced by the 11 heads of the baseline fViT and the part fViT in Fig.~\ref{fig:attention_map}. We observe that for both methods, the heads achieve good correspondence across pose as each head fires at corresponding areas in both the frontal and the profile images. Then, a closer look reveals that the 6-th and 7-th attention heads (6-th and 7-th columns of Fig.~\ref{fig:attention_map}) of the baseline fViT (1-st and 2-nd rows) do not focus on specific facial parts. Moreover, for the baseline fViT there's only one head that focuses on the eyes. This is in stark contrast with the part fViT where there are multiple heads focusing on the eyes region which are well-known to be the most discriminative facial parts for FR~\cite{wang2020hierarchical,xie2018comparator,zhang2012collaborative,lederman2010haptic,schyns2002show,ranjan2017hyperface}. Fig.~\ref{fig:49landmark} shows the 49 landmarks learned by our part fViT. As shown landmark correspondence across pose was learned to some good degree. Besides FR results, our landmark CNN can be useful for providing facial landmarks learned without landmark supervision. The detailed explanation can be observed in the supplementary material Section $2.3$




\section{Conclusions}

We proposed face Transformers as architectures for highly accurate face recognition. We described two models: (a) fViT, our strong baseline trained appropriately on MS1M. (b) part fViT, we capitalized on the Transformer's property to process visual tokens extracted from irregular grids to propose a part-based face Transformer which is trained end-to-end to perform landmark localization and face recognition without explicit landmark supervision. Our pipeline is extremely simple comprising a lightweight CNN for direct coordinate regression followed by a ViT operating on the patches extracted from the predicted landmarks. Both models, and especially our part fViT, achieve state-of-the-art or near state-of-the-art accuracy on several face recognition benchmarks.

\section*{Acknowledgement}
Zhonglin Sun is supported by China Scholarship Council(CSC).

\bibliography{egbib}
\newpage
\appendix

    

\section{Introduction}
This is the supplementary material for the paper \textbf{Part-based Face Recognition with Vision Transformers}. We first exhibit the detailed choice of data augmentation we used to enhance fViT in \textbf{Section}~\ref{training_de}. Then we list the model details adopted for our fViT in \textbf{Section}~\ref{model_details}. Effect of data augmentation, the overlapping rate of landmarks and the comparison of choice of landmark CNN are also included as the additional ablation study in \textbf{Section}~\ref{add_abla}. Finally, we describe learned landmark network is effective for the side task of Application to unsupervised landmark discovery in \textbf{Section}~\ref{landmark_dis}.

\section{Additions to section 4: Experiments}\label{sec:supp_exp}
\subsection{Implementation details}
\subsubsection{Training details}\label{training_de}
For training the Transformer, we opted to use a large amount of data augmentation compared to the original face recognition setting used in ResNets. Specifically, we used stochastic depth regularization with probability 0.1~\cite{larsson2016fractalnet}, resize \& crop in the range $[0.9, 1.0]$, RandAugment~\cite{cubuk2020randaugment} with magnitude of 2, and without the solarize and invert operations, Mixup~\cite{zhang2017mixup} with alpha=0.5 and probability of 0.2, Cutout with value 0.1, and weight decay 1e-1 for the ViT backbone and 5e-2 for the Landmark CNN. We adopted AdamW~\cite{loshchilov2017decoupled} and the cosine learning rate decay followed by warm-up of 5 epochs, while we trained in total for 34 epochs. All networks are trained from scratch. 

\begin{table}[ht]
\centering
\footnotesize
\begin{tabular}{|c|c|c|c|}
\hline
Model & Hidden size & Parameters & FLOPS \\ \hline
part fViT-B & 768 & 66M & 12.64G \\ \hline
fViT-B & 768 & 63M & 12.58G \\ \hline
Resnet-100 & - & 65M & 12.10G \\ \hline \hline
part fViT-S & 512 & 46M & 8.96G \\ \hline
fViT-S & 512 & 43M & 8.90G \\ \hline
Resnet-50 & - & 43.59M & 6.33G \\ \hline
\end{tabular}
\caption{Network sizes and FLOPS for our fViT and Part fViT and Resnet}
\label{model_size}
\end{table}
\subsubsection{Model details}\label{model_details}
To fairly compare with Resnet~\cite{deng2019arcface} which is used as the backbone in most recent methods, we constructed our fViT in order to have a similar model size and FLOPS with Resnet-100. Our base configuration for fViT, called fViT-B, has 12 layers, 11 attention heads and $d=768$. We also built a fViT-S. Our models and Resnet-100 are compared in Table~\ref{model_size}. As can be observed, our fViT-B has similar model size and FLOPS with Resnet-100. Our landmark network is a MobilenetV3~\cite{howard2019searching} unless otherwise specified. All models are implemented in PyTorch~\cite{paszke2017automatic}.

\subsection{Additional Ablation Study}\label{add_abla}

\subsubsection{Effect of different data augmentations}\label{sec:pos_enc}  Here we present effectiveness of the choice of different augmentations suggested in~\cite{steiner2021train} starting from random filp. Results can be found in ~\ref{aug}, we can observe that with more data augmentation are added, more accurate results will be gained.
\begin{table}[h]
\centering
\footnotesize
\scalebox{0.8}{
\begin{tabular}{|c|c|c|c|c|c|c|c|c|c|c|c|}
\hline
Exp & Flip & Randaug & Res\&Crop & Stostich & Mixup & Cutout & Warm-up & LFW & CFP-FP & AgeDB-30 & IJB-C \\ \hline
1 & $\surd$ &   &   &   &   &   &   & 99.63 & 95.72 & 97.1  & 95.29 \\ \hline
2 & $\surd$ & $\surd$ &   &   &   &   &   & 99.68 & 96.84 & 97.55 & 95.87 \\ \hline
3 & $\surd$ & $\surd$ & $\surd$ &   &   &   &   & 99.70 & 97.23 & 97.26 & 95.98 \\ \hline
4 & $\surd$ & $\surd$ & $\surd$ & $\surd$ &   &   &   & 99.73 & 97.40 & 97.30 & 96.05 \\ \hline
5 & $\surd$ & $\surd$ & $\surd$ & $\surd$ & $\surd$ &   &   & 99.76 & 98.19 & 97.60 & 96.13 \\ \hline
6 & $\surd$ & $\surd$ & $\surd$ & $\surd$ & $\surd$ & $\surd$ &   & 99.78 & 98.37 & 97.67 & 96.23 \\ \hline
7 & $\surd$ & $\surd$ & $\surd$ & $\surd$ & $\surd$ & $\surd$ & $\surd$ & 99.80 & 98.78 & 97.85 & 96.37 \\ \hline
\end{tabular}}
\caption{Impact of data augmentation}
\label{aug}
\end{table}

\subsubsection{Degree of overlapping}
We also examiate the degree of overlapping patches trained by our network, we calculate the mean and variance overlap rate of the closest patches, listed in Table~\ref{overlap}.The overlap rate for the large pose datasets CFP-FP\& IJB-C is higher than that for other datasets.
\begin{table}[h]
\centering
\small
\scalebox{1.0}{
\begin{tabular}{|c|c|c|c|c|}
\hline
      & LFW             & CFP-FP         & AgeDB-30        & IJB-C \\ \hline
R=16  & 0.5007\textpm 0.0002  & 0.5250\textpm 0.0016 & 0.4980\textpm 0.0002  & 0.5099\textpm 0.0007      \\ \hline
R=49  & 0.3993\textpm 0.0002 & 0.4665\textpm 0.0064 & 0.3997\textpm 0.0001  &  0.4279\textpm 0.0003     \\ \hline
R=196 & 0.2681\textpm 0.0001  & 0.2950\textpm 0.0010 & 0.2684\textpm 0.00008 & 0.2789\textpm 0.0005      \\ \hline
\end{tabular}}
\caption{The overlap rate of the neighboring patches obtained by our part fVIT-B with R=16, 49 and 196}
\label{overlap}
\end{table}\\

\subsubsection{Effect of different landmark CNNs} We conducted an experiment to evaluate the impact of using different CNNs for landmark network. Specifically, we also chose Resnet-50\cite{he2016deep}. The model used is the part fViT-B, with $R=196$ landmarks. Table~\ref{Resnet} shows the obtained results. We conclude that a larger landmark CNN does not further boost the final accuracy.
\begin{table}[h]
\centering
\small
\begin{tabular}{|c|c|c|c|c|}
\hline
Landmark Network & LFW & CFP-FP & AgeDB & IJB-C  \\ \hline
fViT & 99.85 & 99.01 & 98.13 & 97.21  \\ \hline
part fViT (MobilenetV3) & 99.83 & \textbf{99.21} & 98.29 & \textbf{97.29}  \\ \hline
part fViT (ResNet50) & 99.81 & 99.14 & \textbf{98.35} & 97.11 \\ \hline
\end{tabular}
\caption{Impact of landmark CNNs on face recognition accuracy.}
\label{Resnet}
\end{table}

\subsection{Application to unsupervised landmark discovery}\label{landmark_dis}
We opted for a quantitative evaluation of the facial landmarks discovered by our landmark CNN using the evaluation protocol and codebase of~\cite{sanchez2019object}. Specifically, we follow~\cite{sanchez2019object} and report the so-called forward error on the whole MAFL \& AFLW datasets in Table~\ref{landmark-error}. The forward error is a measure of landmark stability, its pipeline is to train a regressor with predicted landmarks as the training data and 5 manually labelled landmarks on the MAFL \& AFLW datasets as the test set. The more stable the predicted landmarks are, the better they map to the ground truth (for details and forward error definition, please see\cite{sanchez2019object}). As it can be observed our method offers competitive results with recently proposed methods which are exclusively designed for unsupervised landmark localization.  
\begin{table}[h]
\centering
\small
\begin{tabular}{|c|c|c|c|}
\hline
                              & Method           & MAFL          & AFLW          \\ \hline
\multirow{2}{*}{Supervised}   & TCDCN~\cite{zhang2015learning}            & 7.95          & 7.65          \\ \cline{2-4} 
                              & MTCNN~\cite{zhang2014facial}            & 5.39          & 6.90          \\ \hline
\multirow{8}{*}{Unsupervised} & Thewlis~\cite{thewlis2017unsupervised}          & 7.15          & -             \\ \cline{2-4} 
                              & Jakab~\cite{jakab2018unsupervised}          & 3.19 & 6.86          \\ \cline{2-4} 
                              & Zhang~\cite{zhang2018unsupervised}            & 3.46          & 7.01          \\ \cline{2-4} 
                              & Shu~\cite{shu2018deforming}              & 5.45          & -             \\ \cline{2-4} 
                              & Sahasrabudhe~\cite{sahasrabudhe2019lifting}     & 6.07          & -             \\ \cline{2-4} 
                              & Sanchez~\cite{sanchez2019object}          & 3.99          & 6.69 \\ \cline{2-4} 
                              & Mallis~\cite{mallis2020unsupervised}          & 4.12          & 7.37 \\ \cline{2-4} 
                              & Li~\cite{li2021unsupervised}          & \textbf{3.08}         & \textbf{6.20} \\ \cline{2-4} 
                              \hline \hline
\multirow{2}{*}{Ours}         & Landmark CNN  & 4.87          & 10.22          \\ \cline{2-4} 
                              & Landmark CNN ($R=49$)  & 3.37          & 7.16          \\ \cline{2-4} 
                              & Landmark CNN ($R=16$) & 3.88          & 7.69          \\ \hline
\end{tabular}
\caption{Comparison on unsupervised landmark discovery. Forward error results~\cite{sanchez2019object} are reported on the whole MAFL \& AFLW datasets.}
\label{landmark-error}
\end{table}

\end{document}